\documentclass{stars-techreport}

\usepackage{xparse}

\usepackage{amsmath} 
\usepackage{amssymb}
\usepackage{amsthm}
\usepackage{amsfonts} 
\usepackage{color}

\usepackage{graphicx}

\usepackage{hyperref}

\usepackage{cleveref}
\crefformat{equation}{(#2#1#3)}
\Crefname{figure}{Fig.}{Figs.}
\crefname{figure}{Fig.}{Figs.}
\crefformat{figure}{#2Fig.~#1#3}
\Crefformat{figure}{#2Fig.~#1#3}
\crefformat{table}{#2Table~#1#3}
\Crefformat{table}{#2Table~#1#3}
\crefformat{section}{#2Sec.~#1#3}
\Crefformat{section}{#2Sec.~#1#3}
\crefformat{problem}{#2Problem~#1#3}
\Crefformat{problem}{#2Problem~#1#3}
\crefname{appsec}{Appendix}{Appendices}

\usepackage{url}
\usepackage{graphicx}

\usepackage{array}
\usepackage{longtable}
\usepackage{multirow}
\usepackage{float}
\usepackage{caption}
\usepackage{mathrsfs}
\usepackage{subfig}
\usepackage{listings}
\usepackage{rotating}
\usepackage{overpic}

\usepackage{booktabs}

\usepackage{xparse}
\usepackage{tikz}

\newcommand{\lrss}[5]{%
	\setbox1=\hbox{\ensuremath{^{#1}}}%
	\setbox2=\hbox{\ensuremath{_{#2}}}%
	\setbox5=\hbox{\ensuremath{#5}}%
	\setbox6=\hbox{\ensuremath{^{#1#3}}}%
	\setbox7=\hbox{\ensuremath{_{#2#4}}}%
	\setbox8=\hbox{\ensuremath{^{#3}}}%
	\setbox9=\hbox{\ensuremath{_{#4}}}%
	\hspace{\ifnum\wd1>\wd2\wd1\else\wd2\fi}%
	\ensuremath{\copy5%
		^{\hspace{-\wd1}\hspace{\wd1}\hspace{\wd8}%
			\hspace{-\wd6}\hspace{-\wd5}#1\hspace{\wd5}#3}%
		_{\hspace{-\wd2}\hspace{\wd2}\hspace{\wd9}%
			\hspace{-\wd7}\hspace{-\wd5}#2\hspace{\wd5}#4}%
}}

\NewDocumentCommand\bbm{}{ \begin{bmatrix} }
\NewDocumentCommand\ebm{}{ \end{bmatrix} }
\NewDocumentCommand\Vector{m}{ \boldsymbol{\mathbf{#1}} }
\NewDocumentCommand\Matrix{m}{ \boldsymbol{\mathbf{#1}} }
\NewDocumentCommand\UnitVec{m}{ \Vector{\hat{#1}} }

\NewDocumentCommand\Transpose{}{^\mathsf{T}}

\NewDocumentCommand\Vectorize{m}{\mathrm{vec}\left(#1\right)}

\NewDocumentCommand\Real{}{ \mathbb{R} }

\NewDocumentCommand\ExpM{m}{ \Vector{e}^{#1} }

\NewDocumentCommand\NormalDistribution{mm}{ \mathcal{N}\left(#1,#2\right) }

\NewDocumentCommand\CovarianceMatrix{}{ \Matrix{\varSigma} }

\NewDocumentCommand\ZeroMatrix{m}{ \Matrix{0}_{#1\times#1} }
\NewDocumentCommand\IdentityMatrix{m}{ \Matrix{\IdentitySymbol}_{#1\times#1} }

\NewDocumentCommand\CoordinateFrame{m}{ \underrightarrow{\Matrix{\mathcal{F}}}_{#1} }

\NewDocumentCommand\Estimate{m}{\hat{#1}}
\NewDocumentCommand\Prior{m}{\check{#1}}

\NewDocumentCommand\IdentitySymbol{}{ 1 }
\NewDocumentCommand\Skew{m}{\left[#1\right]_\times}

\NewDocumentCommand\KroneckerProduct{}{ \otimes }

\NewDocumentCommand\CrossP{}{ \times }

\NewDocumentCommand\Wrench{}{ \Vector{\mathcal{W}} }
\NewDocumentCommand\Position{}{ \Vector{p} }

\NewDocumentCommand\Acceleration{}{ \Vector{a} }
\NewDocumentCommand\Velocity{}{ \Vector{v} }
\NewDocumentCommand\AngularAcceleration{}{ \Vector{\alpha} }
\NewDocumentCommand\AngularVelocity{}{ \Vector{\omega} }

\NewDocumentCommand\JacobianSymbol{}{ J }
\NewDocumentCommand\Jac{mmm}{ \lrss{#2}{}{#1}{#3}{\Matrix{\JacobianSymbol}}(\GenCoord) }

\NewDocumentCommand\JacobianSpace{}{ \Matrix{\JacobianSymbol}_s(\GenCoord) }
\NewDocumentCommand\HessianSymbol{}{ H }
\NewDocumentCommand\Hess{mmm}{ \lrss{#2}{}{#1}{#3}{\Matrix{\HessianSymbol}}(\GenCoord) }

\NewDocumentCommand\GeneralCoordinates{}{ q }

\NewDocumentCommand\GenCoord{}{ \Vector{\GeneralCoordinates} }

\NewDocumentCommand\RotationSymbol{}{ R }

\NewDocumentCommand\TransformSymbol{}{ T }
\NewDocumentCommand\Rot{mm}{ {\lrss{#2}{}{#1}{}{\Matrix{\RotationSymbol}}\:} }
\NewDocumentCommand\Pos{mmm}{ \lrss{#2}{}{#1}{#3}{\Vector{\Position}}\: }

\NewDocumentCommand\Trans{mmm}{ \lrss{#2}{}{#1}{#3}{\Matrix{\TransformSymbol}}\: }

\NewDocumentCommand\TwistSymbol{}{ \nu }

\NewDocumentCommand\Screw{}{ \Vector{\mathcal{S}} }
\NewDocumentCommand\ScrewAxis{}{ \UnitVec{\Vector{s}} }

\NewDocumentCommand\ScrewExpM{mm}{\ExpM{\Skew{#1}#2}}

\NewDocumentCommand\Adjoint{m}{ \text{Ad}\left(#1\right) }

\NewDocumentCommand\VelTwist{mmm}{ \lrss{#2}{}{#1}{#3}{\mathcal{\Vector{\TwistSymbol}}}\: }
\NewDocumentCommand\LinVel{mmm}{ \lrss{#2}{}{#1}{#3}{\Velocity}\: }
\NewDocumentCommand\AngVel{mmm}{ \lrss{#2}{}{#1}{#3}{\AngularVelocity}\: }

\NewDocumentCommand\LinAcc{mmm}{ \lrss{#2}{}{#1}{#3}{\Acceleration}\: }
\NewDocumentCommand\AngAcc{mmm}{ \lrss{#2}{}{#1}{#3}{\AngularAcceleration}\: }

\RenewDocumentCommand\CoordinateFrame{m}{ \Matrix{\mathcal{F}}_{#1} }
\NewDocumentCommand\CFrame{m}{ \CoordinateFrame{#1} }

\STARStitle{Automated Continuous Force-Torque Sensor Bias Estimation}
\STARSdate{\today}      
\STARSauthor{Philippe Nadeau, Miguel Rogel Garcia, Emmett Wise, and Jonathan Kelly}      
\STARSidentifier{STARS-2024-001}  
\STARSmajorrev{1}                 
\STARSminorrev{2}                

\begin{document}
	\maketitle
	
    \begin{abstract}
    Six axis force-torque sensors are commonly attached to the wrist of serial robots to measure the external forces and torques acting on the robot's end-effector.
    These measurements are used for load identification, contact detection, and human-robot interaction amongst other applications.
    Typically, the measurements obtained from the force-torque sensor are more accurate than estimates computed from joint torque readings, as the former is independent of the robot's dynamic and kinematic models.
    However, the force-torque sensor measurements are affected by a bias that drifts over time, caused by the compounding effects of temperature changes, mechanical stresses, and other factors \cite{kroger200812d}.
    In this work, we present a pipeline that continuously estimates the bias and the drift of the bias of a force-torque sensor attached to the wrist of a robot.
    The first component of the pipeline is a Kalman filter that estimates the kinematic state (position, velocity, and acceleration) of the robot's joints.
    The second component is a kinematic model that maps the joint-space kinematics to the task-space kinematics of the force-torque sensor.
    Finally, the third component is a Kalman filter that estimates the bias and the drift of the bias of the force-torque sensor assuming that the inertial parameters of the gripper attached to the distal end of the force-torque sensor are known with certainty.
    \end{abstract}

    \section{Manipulator Joint's State Estimation}
    We consider three distinct models for the estimation of the joint's state. 
    The first model assumes that noise is only present in the joint's acceleration and does not impact the joint's position and velocity estimates.
    The second model assumes that noise is present in the joint's position, velocity, and acceleration, to consider the impact of the uncertainty in the computation of $\Delta t$, the time interval between two consecutive measurements.
    In practice, $\Delta t$ could either be obtained by computing the difference between data packet arrival time or be made constant by fixing the estimation process to be done at specific time intervals. 
    Our third model, which is the one that is actually used in our proposed pipeline, goes one step further and assumes that noise is caused by the jerk of the joint, and integrated such that the position, velocity and acceleration estimates are affected by the noise.

    %
    In the following, the state $\Vector{q}_j=[q_j, \dot{q}_j, \ddot{q}_j] \in \Real^3$ of each joint $j$ is estimated with respect to the previous link; in other words, it is estimated about the joint's own axis of rotation. 
    For conciseness, the subscript $j$ is omitted from the equations.
    The motion model is given by

    \begin{align}
        \Vector{q}_{i+1}&=\begin{bmatrix}
            q \\
            \dot{q} \\
            \ddot{q}
        \end{bmatrix}
        =\underbrace{\begin{bmatrix}
            1 & \Delta t & 0 \\
            0 & 1 & \Delta t \\
            0 & 0 & 1 \\
        \end{bmatrix}}_{\Matrix{A}_i}
        \Vector{q}_i
        + \Vector{m}_i, \quad \Vector{m}_i \sim \mathcal{N}(\Vector{0}, \underbrace{[\sigma^2_q, \sigma^2_{\dot{q}}, \sigma^2_{\ddot{q}}]^T\IdentityMatrix{3}}_{\Matrix{Q}_i})~,
    \end{align}
    where the process noise $\Vector{m}_i$ affects all three states and is assumed to be Gaussian. 
    
    \subsection{Non-Integrated Acceleration Noise}
    In \cite{kubus2007line}, it is assumed that noise is only affecting the joint acceleration component $\ddot{q}$ such that $\sigma_q$, $\sigma_{\dot{q}} = 0$. 
    
    Assuming that the robot is initially stationary, the initial condition is given by
    \begin{align*}
        \Vector{q}_0 = \bbm
        q_0 \\ 
        0 \\
        0
        \ebm~,
    \end{align*}
    where $q_0$ is the measured initial joint position.
    
    The observation model for the estimation of joint states is given by
    \begin{align}
        \Vector{y}_{i}=\begin{bmatrix}
            \bar{q} \\
            \dot{\bar{q}}
        \end{bmatrix}
        =\underbrace{\begin{bmatrix}
            1 & 0 & 0 \\
            0 & 1 & 0 \\
        \end{bmatrix}}_{\Matrix{C}_i}
        \Vector{q}_i
        + \Vector{o}_i \quad \Vector{o}_i \sim \mathcal{N}(\Vector{0}, \underbrace{[\eta^2_q, \eta^2_{\dot{q}}]^T\IdentityMatrix{2}}_{\Matrix{R}_i})~,
    \end{align}
    which yields the measurement vector $\mathbf{y}_i$, where the noise $\Vector{o}_i$ on those measurements is assumed to be Gaussian.
    
    \subsection{Constant Acceleration with Integrated Acceleration Noise}
    \label{sec:integrated_accel_noise}
    We can derive an alternative discrete motion model for the joint's state of the robot by following example 3.2 of \cite{barfoot2024state}. 
    This assumes that we have a continuous time model with constant acceleration. 
    The state to be estimated becomes
    \begin{align}
        \Vector{q}(t)        &= \bbm q(t) & \dot{q}(t) & \ddot{q}(t) \ebm^T.
    \end{align}
    Assuming that white noise enters the model through the acceleration, we have
    \begin{align}
        \ddot{q}(t) &= \textrm{w}(t), \quad \textrm{w}(t) \sim \mathcal{GP}(0, \sigma^2 \delta(t-t')),
    \end{align}
    where $\sigma^2=\mathbf{Q}\in\Real$. 
    The continuous time model is thus
    \begin{align}
        \label{eq:joint_continuous}
        \dot{\Vector{q}}(t) = \Matrix{A}\Vector{q}(t) + \Matrix{L}\textrm{w}(t),
    \end{align}
    where the matrices in the model are respectively defined as
    \begin{align}
        \Matrix{A} = \bbm 0 & 1 & 0 \\ 0 & 0 & 1 \\ 0 & 0 & 0 \ebm, \quad 
        \Matrix{L} = \bbm 0 \\ 1 \\ 0 \ebm~.
    \end{align}
    Taking the exponential, we obtain the following transition function
    \begin{align}
        \exp\left({\Matrix{A}\Delta t}\right) & = \IdentityMatrix{3} + \Matrix{A}\Delta t + \frac{1}{2}\Matrix{A}^2\Delta t^2 + \frac{1}{6}\underbrace{\Matrix{A}^3}_{\ZeroMatrix{3}}\Delta t^3 + \ldots \\
        & = \bbm 1 & \Delta t & \frac{1}{2} \Delta t^2  \\
                 0 &     1    &  \Delta t   \\
                 0 &     0    &     1\ebm.
    \end{align}
    Following \cite{barfoot2024state}, $\Matrix{Q}_i$ for the discrete time process is computed as
    \begin{align}
        &\Matrix{Q_i} = \int_{0}^{\Delta t_{k:k-1}}  
        \exp\left({\Matrix{A}(\Delta t_{k:k-1}-s)}\right) 
        \Matrix{L} \Matrix{Q} \Matrix{L}^T 
        \exp\left({\Matrix{A}(\Delta t_{k:k-1}-s)}\right)^T \,ds \nonumber \\
        & = \bbm \frac{1}{3}\Delta t_{k:k-1}^3 \Matrix{Q}& \frac{1}{2}\Delta t_{k:k-1}^2 \Matrix{Q} & 0 \\
        \frac{1}{2}\Delta t_{k:k-1}^2 \Matrix{Q} & \Delta t_{k:k-1}\Matrix{Q} &  0 \\
        0 & 0    &  0\ebm
    \end{align}
    which has no uncertainty associated with the acceleration term of the discrete time process, as expected.

    \subsection{Constant Acceleration with Integrated Jerk Noise}
    Starting from the model outlined in \cref{sec:integrated_accel_noise}, but assuming that white noise enters the model through the jerk, we obtain
    \begin{align}
        \dddot{q}(t)        &= \textrm{w}(t), \quad \textrm{w}(t) \sim \mathcal{GP}(0, \sigma^2 \delta(t-t')),
    \end{align}
    where $\sigma^2=\mathbf{Q}\in\Real$.
    The matrices of the continuous model in \cref{eq:joint_continuous} become
    \begin{align}
        \Matrix{A} = \bbm 0 & 1 & 0 \\ 0 & 0 & 1 \\ 0 & 0 & 0 \ebm, \quad 
        \Matrix{L} = \bbm 0 \\ 0 \\ 1 \ebm.
    \end{align}
    The covariance matrix $\Matrix{Q}_i$ for the discrete time process can be computed according to \cite{barfoot2024state} with
    \begin{align}
        &\Matrix{Q_i} = \int_{0}^{\Delta t_{k:k-1}}  
        \exp\left({\Matrix{A}(\Delta t_{k:k-1}-s)}\right) 
        \Matrix{L} \Matrix{Q} \Matrix{L}^T 
        \exp\left({\Matrix{A}(\Delta t_{k:k-1}-s)}\right)^T \,ds \nonumber \\
        & = \int_{0}^{\Delta t_{k:k-1}}  
        \bbm 1 & \Delta t_{k:k-1}-s & \frac{1}{2} (\Delta t_{k:k-1}-s)^2  \\
                 0 &     1    &  \Delta t_{k:k-1}-s   \\
                 0 &     0    &     1\ebm
        \bbm 0 \\ 0 \\ 1 \ebm \Matrix{Q} \bbm 0 & 0 & 1 \ebm
        \bbm 1 & 0 & 0                      \\
        \Delta t_{k:k-1}-s &     1    &  0   \\
        \frac{1}{2}(\Delta t_{k:k-1}-s)^2 &     \Delta t_{k:k-1}-s    &     1\ebm \,ds \nonumber\\
        & = \int_{0}^{\Delta t_{k:k-1}} 
        \bbm \frac{1}{4}(\Delta t_{k:k-1}-s)^4 \Matrix{Q}& \frac{1}{2}(\Delta t_{k:k-1}-s)^3 \Matrix{Q} & \frac{1}{2}(\Delta t_{k:k-1}-s)^2 \Matrix{Q} \\
        \frac{1}{2}(\Delta t_{k:k-1}-s)^3 \Matrix{Q} & (\Delta t_{k:k-1}-s)^2 \Matrix{Q} &  (\Delta t_{k:k-1}-s)  \Matrix{Q} \\
        \frac{1}{2}(\Delta t_{k:k-1}-s)^2 \Matrix{Q} &     (\Delta t_{k:k-1}-s) \Matrix{Q}    &     \Matrix{Q}\ebm \,ds \nonumber \\
        & = \bbm \frac{1}{20}\Delta t_{k:k-1}^5 \Matrix{Q}& \frac{1}{8}\Delta t_{k:k-1}^4 \Matrix{Q} & \frac{1}{6}\Delta t_{k:k-1}^3 \Matrix{Q} \\
        \frac{1}{8}\Delta t_{k:k-1}^4 \Matrix{Q} & \frac{1}{3}\Delta t_{k:k-1}^3 \Matrix{Q} &  \frac{1}{2}\Delta t_{k:k-1}^2  \Matrix{Q} \\
        \frac{1}{6}\Delta t_{k:k-1}^3 \Matrix{Q} & \frac{1}{2}\Delta t_{k:k-1}^2 \Matrix{Q}    &     \Delta t_{k:k-1}\Matrix{Q}\ebm,
    \end{align}
    such that the discrete time model becomes 
    \begin{align}
        \label{eq:joint_discrete}
        \Vector{q}_{i+1}&=
        \underbrace{\begin{bmatrix}
            1 & \Delta t & \frac{1}{2} \Delta t^2 \\
            0 & 1 & \Delta t \\
            0 & 0 & 1 \\
        \end{bmatrix}}_{\Matrix{A}_i}
        \Vector{q}_i
        + \Vector{m}_i, \quad \Vector{m}_i \sim \mathcal{N}(\Vector{0}, \Matrix{Q}_i).
    \end{align}
    The assumption of white noise jerk is made such that jerk will be minimized as much as possible in the optimization process, acting similarly to a regularizer favoring constant acceleration. 
    
    \section{Force-Torque Sensor Kinematics}
    Since wrench measurements are \textit{observed} in the force-torque sensor frame, the kinematics of the sensor is ultimately what matters.
    To map joint-space kinematics to task-space kinematics, the forward kinematics must be carried and the Jacobian must be evaluated at each time step.
    
    Let $\Screw_i$ be the definition of the unit screw axis of the $i$-th joint expressed relative to the base of the robot $\CFrame{w}$ when the configuration of the robot is $\Vector{q} = \Vector{0}$.
    With $\Screw_i = \bbm -\ScrewAxis_i \CrossP \Vector{r}_i & \ScrewAxis_i \ebm\Transpose$, where $\ScrewAxis$ is the direction of the screw axis and $\Vector{r}$ is a point on the screw axis, the screw axes for the Franka Panda and Franka Research 3 (FR3) robots are given by
    \begin{align}
        \Screw_1 &= \bbm 0 & 0 & 0 & 0 & 0 & 1 \ebm\Transpose\\&\text{ as }
        \ScrewAxis_1 = \bbm 0 & 0 & 1 \ebm\Transpose \text{ and }
        \Vector{r}_1 = \bbm 0 & 0 & 0 \ebm\Transpose\notag\\
        \Screw_2 &= \bbm -0.333 & 0 & 0 & 0 & 1 & 0 \ebm\Transpose\\&\text{ as }
        \ScrewAxis_2 = \bbm 0 & 1 & 0 \ebm\Transpose \text{ and }
        \Vector{r}_2 = \bbm 0 & 0 & 0.333 \ebm\Transpose\notag\\
        \Screw_3 &= \bbm 0 & 0 & 0 & 0 & 0 & 1 \ebm\Transpose\\&\text{ as }
        \ScrewAxis_3 = \bbm 0 & 0 & 1 \ebm\Transpose \text{ and }
        \Vector{r}_3 = \bbm 0 & 0 & 0.649 \ebm\Transpose\notag\\
        \Screw_4 &= \bbm 0.649 & 0 & -0.0825 & 0 & -1 & 0 \ebm\Transpose\\&\text{ as }
        \ScrewAxis_4 = \bbm 0 & -1 & 0 \ebm\Transpose \text{ and }
        \Vector{r}_4 = \bbm 0.0825 & 0 & 0.649 \ebm\Transpose\notag\\
        \Screw_5 &= \bbm 0 & 0 & 0 & 0 & 0 & 1 \ebm\Transpose\\&\text{ as }
        \ScrewAxis_5 = \bbm 0 & 0 & 1 \ebm\Transpose \text{ and }
        \Vector{r}_5 = \bbm 0 & 0 & 1.033 \ebm\Transpose\notag\\
        \Screw_6 &= \bbm 1.033 & 0 & 0 & 0 & -1 & 0 \ebm\Transpose\\&\text{ as }
        \ScrewAxis_6 = \bbm 0 & -1 & 0 \ebm\Transpose \text{ and }
        \Vector{r}_6 = \bbm 0 & 0 & 1.033 \ebm\Transpose\notag\\
        \Screw_7 &= \bbm 0 & 0.088 & 0 & 0 & 0 & -1 \ebm\Transpose\\&\text{ as }
        \ScrewAxis_7 = \bbm 0 & 0 & -1 \ebm\Transpose \text{ and }
        \Vector{r}_7 = \bbm 0.088 & 0 & 0.926 \ebm\Transpose\notag
    \end{align}
    where the values are the nominal values from the model of the robot.

    The Jacobian $\JacobianSpace$ is built column by column, where the $j$-th column $\JacobianSpace_j$ is the screw axis $\Screw_j$ of the corresponding joint.
    The screw axis is expressed in the base frame $\CFrame{w}$ through the adjoint transformation $\Adjoint{\Trans{j}{w}{w}}$.
    In parallel, the forward kinematics is computed to obtain the pose of the sensor frame $\CFrame{s}$ with respect to the base frame $\CFrame{w}$.
    Iterating over $j = 1, \ldots, N_j$, where $N_j$ is the number of joints, compute 
    \begin{align}
        \JacobianSpace_j &= 
        \underset{\Adjoint{\Trans{j}{w}{w}}}{\underbrace{
            \bbm
            \Rot{j}{w} & \Skew{\Pos{j}{w}{w}}\Rot{j}{w}\\
            \Matrix{0}_{3\times 3} & \Rot{j}{w}
            \ebm
        }}
        \Screw_j\\
        \Trans{j}{w}{w} &= \Trans{j-1}{w}{w} \ScrewExpM{\Screw_j}{q_j}
    \end{align}
    where $\ScrewExpM{\Screw_j}{q_j}$ is the exponential map that maps twists to rigid body transformations. Since the screw is defined with respect to the base frame, it is composed through post-multiplication.

    Most computations involved in computing the exponential map can be cached on a per-joint basis, such as to avoid recomputation of the same terms at each time step.
    The exponential mapping is given by
    \begin{align}
        \Rot{j}{w} &= \ScrewExpM{\ScrewAxis_j}{q_j}\\
        &= \IdentityMatrix{3} + \sin(q_j)\Skew{\ScrewAxis_j} + (1 - \cos(q_j))\Skew{\ScrewAxis_j}^2\\
        \Pos{j}{w}{w} &= \left(\IdentityMatrix{3}q_j + \left(1 - \cos(q_j)\right)\Skew{\ScrewAxis_j} + \left(q_j - \sin(q_j)\right)\Skew{\ScrewAxis_j}^2\right)\left(-\Skew{\ScrewAxis_j}\Vector{r}_j\right)\\
        \Trans{j}{w}{w} &= \bbm \Rot{j}{w} & \Pos{j}{w}{w} \\ \Matrix{0}_{1\times 3} & 1 \ebm
    \end{align}
    where the screw definition $\bbm \ScrewAxis & \Vector{r} \ebm$ does not change over time.

    Once the Jacobian is computed, the pose of the sensor is obtained with 
    \begin{align}
        \Trans{s}{w}{w} &= \Trans{N_j}{w}{w} \Trans{s}{w}{w}\vert_{\Vector{q} = \Vector{0}}
    \end{align}
    where $\Trans{s}{w}{w}\vert_{\Vector{q} = \Vector{0}}$ is the nominal pose of the sensor frame with respect to the base frame when the robot is in the zero configuration.
    With the velocity twist of the sensor frame given by $\VelTwist{s}{w}{w} = \JacobianSpace \Vector{\Dot{q}}$, the linear and angular velocities of the sensor frame are given by
    \begin{align}
        \bbm \LinVel{s}{w}{w} \\ \AngVel{s}{w}{w} \ebm &= \underset{\Jac{}{\nu}{s}}{\underbrace{\bbm \IdentityMatrix{3} & -\Skew{\Pos{s}{w}{w}} \\ \Matrix{0}_{3\times 3} & \IdentityMatrix{3} \ebm \JacobianSpace} } \Vector{\Dot{q}}
    \end{align}
    where $\LinVel{s}{w}{w}$ and $\AngVel{s}{w}{w}$ are the linear and angular velocities of the sensor frame with respect to the base frame, respectively, that are used to compute the data matrix $\Matrix{D}_i$.
    The Jacobian $\Jac{}{\nu}{s}$ maps joint velocities to sensor frame linear and angular velocities.

    The acceleration of the sensor frame is given by 
    \begin{align}
        \bbm \LinAcc{s}{w}{w} \\ \AngAcc{s}{w}{w} \ebm &= \left(\Hess{}{\nu}{s} \Vector{\Dot{q}}\right)\Vector{\Dot{q}} + \Jac{}{\nu}{s} \Vector{\Ddot{q}}
    \end{align}
    where $\Hess{}{\nu}{s}$ is the Hessian, the partial derivative tensor of the Jacobian with respect to the joint coordinates.
    The Hessian of size $\Real^{N_j} \times \Real^6 \times \Real^{N_j}$ can be computed from $\Jac{}{\nu}{s}$ with 
    \begin{align}
        \Hess{}{\nu}{s}_{j,0:6,k} = \Hess{}{\nu}{s}_{k,0:6,j} = \bbm 
        \Skew{\Jac{}{\nu}{s}_{3:6,j}}\Jac{}{\nu}{s}_{0:3,k}\\
        \Skew{\Jac{}{\nu}{s}_{3:6,j}}\Jac{}{\nu}{s}_{3:6,k}
        \ebm
    \end{align}
    where $\Matrix{A}_{a:b,c}$ is the vector built from elements from row $a$ to row $b$ in column $c$ of matrix $\Matrix{A}$, and $\Hess{}{\nu}{s}_{j,0:6,k}$ is the $k$-th column of the $j$-th slice of the Hessian tensor. Computation of the Hessian using this approach has a time complexity of $O(N_j^2)$, which is acceptable for standard serial robots \cite{haviland_2023_mandiff2}.

    \section{Force-Torque Sensor Bias Estimation}
    \label{sec:ft_bias_estimation}
    The objective is to estimate the bias and the slope at which it drifts, assuming that the drift does not accelerate.
    Hence, with $\Vector{b}$ being the bias, $\Vector{\Dot{b}}$ being the drift, then it is assumed that $\Vector{\Ddot{b}} = \Vector{0}$.

    Let,
    \begin{align}
        \Vector{x}_i        &= \bbm \Vector{b}_i & \Vector{\Dot{b}}_i \ebm^T\\
        \Vector{\Dot{x}}_i  &= \bbm \Vector{\Dot{b}}_i & 0 \ebm^T
    \end{align}
    be the state and the derivative of the state, respectively, at the $i$-th time step.

    The process model is given by,
    \begin{equation}
        \Matrix{A}_i = \bbm \IdentityMatrix{6} & \Delta t_i \IdentityMatrix{6} \\ \Matrix{0}_{6\times 6} & \IdentityMatrix{6} \ebm
    \end{equation}
    such that,
    \begin{equation}
        \Vector{x}_{i+1} = \Matrix{A}_i \Vector{x}_i + \Vector{n}_m
    \end{equation}
    where
    \begin{equation}
        \Vector{n}_m \sim \NormalDistribution{\Vector{0}}{\Matrix{Q}_i}
    \end{equation}
    is the uncertainty in the bias estimate at the $i$-th timestep, given by
    \begin{align}
        \label{eq:bias_uncertainty_i}
        \Matrix{Q}_i = \bbm \frac{1}{3} \Delta t_i^3 \Matrix{Q} & \frac{1}{2} \Delta t_i^2 \Matrix{Q} \\ \frac{1}{2} \Delta t_i^2 \Matrix{Q} & \Delta t_i \Matrix{Q} \ebm
    \end{align}
    where $\Matrix{Q} = \sigma_Q^2\IdentityMatrix{6}$ is the process noise covariance matrix and $\Delta t_i = t_{i+1} - t_i$ is the measurement time interval \cite{barfoot2024state}.

    Making use of the identity
    \begin{equation}
        \Vectorize{\Matrix{X}\Matrix{Y}\Matrix{Z}} = \left(\Matrix{Z}\Transpose\KroneckerProduct\Matrix{X}\right)\Vectorize{\Matrix{Y}}
    \end{equation}
    the measurement model is given by
    \begin{align}
        \Wrench_i + \Vector{b}_i + \Vector{n}_w &= \left(\Matrix{D_i} + \Vector{n}_D \right) \Vector{\theta}\\
        &= \IdentityMatrix{6}\left(\Matrix{D_i} + \Vector{n}_D \right) \Vector{\theta}\\
        &= \left(\Vector{\theta}\Transpose \KroneckerProduct \IdentityMatrix{6}\right) \left(\Vectorize{\Matrix{D}_i} + \Vector{n}_D \right)\\
        &= \left(\Vector{\theta}\Transpose \KroneckerProduct \IdentityMatrix{6}\right) \Vectorize{\Matrix{D}_i} + \left(\Vector{\theta}\Transpose \KroneckerProduct \IdentityMatrix{6}\right)\Vector{n}_D
        \label{eq:measurement_model}
    \end{align}
    where $\Vectorize{}$ is the vectorization operator, $\KroneckerProduct$ is the Kronecker product, $\Wrench_i$ is the measured wrench vector ($6\times 1$), $\Matrix{D}_i$ is the data matrix ($6\times 10$), $\Vector{\theta}$ is the known vector of the load's inertial parameters ($10\times 1$), $\Vector{n}_D\sim \NormalDistribution{\Vector{0}}{\CovarianceMatrix_D}$ is the uncertainty in the data matrix, and $\Vector{n}_w\sim \NormalDistribution{\Vector{0}}{\CovarianceMatrix_w}$ is the noise in the wrench measurement.

    The data matrix $\Matrix{D}_i$ is defined by
    \begin{align}
        \Matrix{D}_i &=
        \bbm
            \LinAcc{s_i}{w}{}      & \Matrix{0}_{3\times 9}\\
            \Matrix{0}_{3\times 1} & \Matrix{0}_{3\times 9}
        \ebm
        \\\notag&+
        \bbm
            \Matrix{0}_{3\times 1} & \Skew{\AngVel{s_i}{w}{}}\Skew{\AngVel{s_i}{w}{}}+\Skew{\AngAcc{s_i}{w}{}} & \Matrix{0}_{3\times 6}\\
            \Matrix{0}_{3\times 1} & -\Skew{\LinAcc{s_i}{w}{}} & \Matrix{0}_{3\times 6}
        \ebm
       \\\notag&+
       \bbm
           &&&\Matrix{0}_{3\times 10}&&&\\
           \Matrix{0}_{1\times 4} & \alpha_{x} & \alpha_{y}-\omega_{x}\omega_{z} & \alpha_{z}+\omega_{x}\omega_{y} & -\omega_{y}\omega_{z} & \omega^2_{y}-\omega^2_{z} & \omega_{y}\omega_{z}\\
           \Matrix{0}_{1\times 4} & \omega_{x}\omega_{z} & \alpha_{x}+\omega_{y}\omega_{z} & \omega^2_{z}-\omega^2_{x} & \alpha_{y} & \alpha_{z}-\omega_{x}\omega_{y} & -\omega_{x}\omega_{z}\\
           \Matrix{0}_{1\times 4} & -\omega_{x}\omega_{y} & \omega^2_{x}-\omega^2_{y} & \alpha_{x}-\omega_{y}\omega_{z} & \omega_{x}\omega_{y} & \alpha_{y}+\omega_{x}\omega_{z} & \alpha_{z}
       \ebm 	
    \end{align}
    where $\LinAcc{s_i}{w}{}$, $\AngVel{s_i}{w}{}$, and $\AngAcc{s_i}{w}{}$ are the linear acceleration, angular velocity, and angular acceleration of the sensor frame $s_i$ with respect to the world frame $w$, respectively, such that
    \begin{equation}
        \Wrench_i = \Matrix{D}_i \Vector{\theta} 
    \end{equation}
    when noise is not present, with $\Skew{\cdot}$ being the skew-symmetric matrix operator.

    Since each element of $\Wrench$ depends only on a selection of the elements in $\Vector{\theta}$, a total of $24$ elements of $\Matrix{D}_i$ are always zero.
    Estimation can be simplified by removing the zero elements from $\Vectorize{\Matrix{D}_i}$ and the corresponding columns from $\left(\Vector{\theta}\Transpose \KroneckerProduct \IdentityMatrix{6}\right)$ in \cref{eq:measurement_model} such that $\Vector{n}_D$ is smaller (i.e., $36\times 1$ instead of $60\times 1$).

    To simplify notation, let
    \begin{equation}
        \Matrix{B} = \Vector{\theta}\Transpose \KroneckerProduct \IdentityMatrix{6}
    \end{equation}
    such that
    \begin{align}
        \Wrench_i + \Vector{b}_i + \Vector{n}_w &= \Matrix{B} \Vectorize{\Matrix{D}_i} + \Matrix{B}\Vector{n}_D\\
        \Matrix{B} \Vectorize{\Matrix{D}_i} - \Wrench_i &= \Vector{b}_i + \Matrix{B}\Vector{n}_D + \Vector{n}_w
    \end{align}
    is the measurement model.

    The observation model is given by
    \begin{align}
        \Vector{y}_i &= \Matrix{C} \Vector{x}_i + \Vector{n}_y\\
        &= \Vector{b}_i + \Vector{n}_y\\
        &= \Matrix{B} \Vectorize{\Matrix{D}_i} - \Wrench_i + \Matrix{B}\Vector{n}_D + \Vector{n}_y
    \end{align}
    where $\Matrix{C} = \bbm \Vector{\IdentitySymbol}_{6\times 6} & \Matrix{0}_{6\times 6} \ebm$, $\Vector{n}_y\sim \NormalDistribution{\Vector{0}}{\Matrix{R}_i}$, and $\Matrix{R}_i = \Matrix{B}\CovarianceMatrix_D\Matrix{B}\Transpose + \CovarianceMatrix_W$.

    Setting $\Prior{\Matrix{P}}_0$ as a fairly large diagonal matrix, and $\Prior{\Vector{x}}_0$ as a previously estimated bias or as a zero vector, the prediction step of the Kalman filter is given by
    \begin{align}
        \Prior{\Vector{x}}_{i} &= \Matrix{A}_{i-1} \Estimate{\Vector{x}}_{i-1}\\
        \Prior{\Matrix{P}}_{i} &= \Matrix{A}_{i-1} \Estimate{\Matrix{P}}_{i-1} \Matrix{A}_{i-1}\Transpose + \Matrix{Q}_{i}
    \end{align}
    and the Kalman gain is given by
    \begin{equation}
        \Matrix{K}_{i} = \Prior{\Matrix{P}}_{i} \Matrix{C}\Transpose \left(\Matrix{C} \Prior{\Matrix{P}}_{i} \Matrix{C}\Transpose + \Matrix{R}_{i}\right)^{-1}
    \end{equation}
    such that the correction step is given by
    \begin{align}
        \label{eq:kalman_mean_correction}
        \Estimate{\Vector{x}}_{i} &= \Prior{\Vector{x}}_{i} + \Matrix{K}_{i} \left(\Vector{y}_{i} - \Matrix{C} \Prior{\Vector{x}}_{i}\right)\\
        \Estimate{\Matrix{P}}_{i} &= \left(\IdentityMatrix{12} - \Matrix{K}_{i} \Matrix{C}\right) \Prior{\Matrix{P}}_{i}
    \end{align}
    where $\Estimate{\Vector{x}}_{i}$ is the estimated bias and drift, and $\Estimate{\Matrix{P}}_{i}$ is the associated covariance.
    A biased wrench measurement can be corrected with
    \begin{equation}
        \Wrench_{unbiased} = \Wrench_{biased} + \bbm \Vector{\IdentitySymbol}_{6\times 6} & \Vector{\IdentitySymbol}_{6\times 6}\Delta T\ebm\Estimate{\Vector{x}}_{i}
    \end{equation}
    where $\Delta T$ is the time elapsed since $\Estimate{\Vector{x}}_{i}$ was computed.

    \bibliographystyle{plain}
    \bibliography{references.bib}

    \end{document}